\documentclass[acmtog,nonacm]{acmart}
\acmSubmissionID{1860}

\usepackage{multirow} 

\usepackage{graphicx}

\usepackage[ruled]{algorithm2e} 

\SetAlFnt{\small}
\SetAlCapFnt{\small}
\SetAlCapNameFnt{\small}
\SetAlCapHSkip{0pt}

\usepackage{xcolor}
\usepackage{caption}
\usepackage[table]{xcolor}
\usepackage{marvosym}
\usepackage{float} 

\usepackage{subcaption}

\acmJournal{TOG}

\begin{document}
\title{ObjFiller3D: Scaling 3D Object Inpainting to Dense Multi-View Consistency}

\author{Haitang Feng}
\authornote{Equal contribution.}
\affiliation{%
  \institution{Nanjing University}
  \country{China}}

\author{Xinkai Chen}
\authornotemark[1]
\affiliation{%
  \institution{Great Bay University}
  \country{China}}

\author{Jie Liu}
\authornote{Corresponding authors.}
\affiliation{%
  \institution{Nanjing University}
  \country{China}}

\author{Jie Tang}
\affiliation{%
  \institution{Nanjing University}
  \country{China}}

\author{Gangshan Wu}
\affiliation{%
  \institution{Nanjing University}
  \country{China}}

\author{Beiqi Chen}
\affiliation{%
  \institution{Harbin Institute of Technology}
  \country{China}}

\author{Jianhuang Lai}
\affiliation{%
  \institution{Sun Yat-sen University}
  \country{China}}

\author{Guangcong Wang}
\authornotemark[2]
\affiliation{%
  \institution{Great Bay University}
  \country{China}}


\renewcommand\shortauthors{Feng, Chen, et al.}

\begin{abstract}
3D object inpainting is commonly achieved via multi-view 2D image completion, yet independently inpainted views often suffer from cross-view inconsistencies, leading to blurred textures, geometric discontinuities, and visual artifacts in the reconstructed 3D objects. To overcome these limitations, we propose ObjFiller-3D, a novel method designed for the completion and editing of high-quality and consistent 3D objects. Instead of relying on sparse-view editing or per-view 2D inpainting, our method jointly optimizes a sequence of densely sampled views along a $360^\circ$ trajectory, enabling global coherence across viewpoints. We design a new framework with three complementary components: a Temporal-Driven Generative Encoder for modeling dense-view dependencies, a Semantic-Aware Completion Encoder for object-level inpainting, and a Cycle-Consistent 3D Encoder that enforces global coherence through a closed-loop formulation. 
Our framework also supports reference-guided 3D inpainting, allowing fine-grained control over appearance. Extensive experiments on diverse datasets demonstrate that ObjFiller-3D significantly outperforms prior methods, achieving higher reconstruction fidelity (PSNR 26.6 vs.\ 15.9 of NeRFiller) and perceptual quality (LPIPS 0.19 vs.\ 0.25 of Instant3dit), while reducing reconstruction time from over 40 minutes to under 10 minutes. These results highlight the effectiveness and practical potential of our approach for real-world 3D editing applications.
\end{abstract}

\begin{teaserfigure}
  \includegraphics[width=\textwidth]{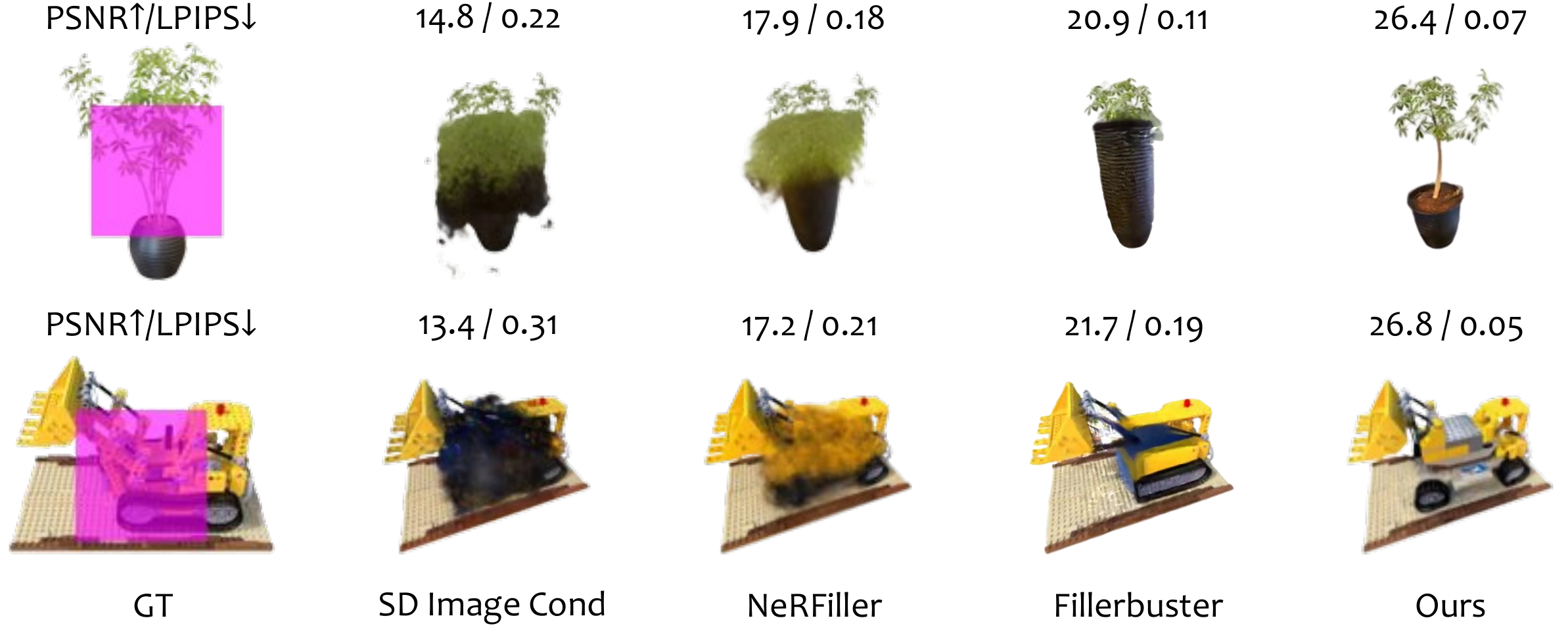}
  \caption{ObjFiller3D reconstructs complete 3D objects from partial inputs. Regions requiring inpainting are marked in pink. Our method outperforms previous state-of-the-art methods across multiple benchmark datasets.}
  \Description{A teaser figure comparing partial 3D object inputs and the completed results produced by ObjFiller-3D, with missing regions highlighted in pink.}
  \label{fig:demo1}
\end{teaserfigure}

\maketitle

\section{Introduction}

High-quality completion of incomplete 3D objects is essential for many graphics applications, including digital reconstruction, 3D asset creation, and game or film production. Environmental constraints during scanning often result in data loss, which poses significant challenges in recovering the original geometry of objects. Despite its importance, research on 3D object-centric completion remains relatively scarce, with only a few methods \cite{weber2024nerfiller,barda2025instant3dit,weber2025fillerbuster,li2025voxhammer} have focused on this problem, and the task is fundamentally distinct from 3D generation and mainstream editing tasks.

Some existing methods offer a partial solution to 3D object inpainting. 3D generation methods~\cite{poole2022dreamfusion, wang2023imagedream, liu2023syncdreamer, long2024wonder3d, gao2024cat3d, hong2023lrm, li2023instant3d, xu2024instantmesh, li2024craftsman3d, xiang2025structured, wu2025direct3d} typically aim to produce high-quality 3D assets from noise, a single image, or a few reference views. These approaches generally assume that the input images depict complete views of the object and are not designed to handle partial or degraded observations. On the other hand, mainstream 3D editing~\cite{haque2023instruct, wang2024view, chen2024gaussianeditor, chen2024proedit, chen2024dge, mirzaei2023spin, liu2024infusion, wu2025aurafusion360, shi2025imfine} focuses on modifying global textures or performing style transfer, without accommodating significant geometric alterations, which makes it unsuitable for restoring missing object parts.

One straightforward approach is to first employ a 2D inpainting model~\cite{suvorov2022resolution,lugmayr2022repaint} to restore multi-view images of an incomplete object. Score Distillation Sampling (SDS)~\cite{poole2022dreamfusion} uses a pre-trained 2D diffusion model as a prior to guide 3D optimization. It renders images from a 3D representation under random viewpoints and backpropagates score-based gradients from the diffusion model to iteratively update the 3D parameters toward the given condition. In contrast, Iterative Dataset Update (IDU)~\cite{haque2023instruct} follows a data-centric paradigm that alternates between 2D image editing and 3D reconstruction. It iteratively refines input images using a generative model and reconstructs the 3D structure from the updated images, progressively improving alignment with the target. However, these methods lack explicit modeling of cross-view dependencies of the same object, which often results in inconsistencies across different viewpoints (see Fig.~\ref{fig:demo1}), which in turn introduces visual artifacts and degrades reconstruction accuracy. Moreover, they do not support high-quality localized 3D editing.

To learn cross-view-aware consistency, several sparse-view editing methods have been proposed. NeRFiller~\cite{weber2024nerfiller} extends IDU by introducing a $2 \times 2$ image grid that packs four views into a single image. It shows that jointly denoising these four views leads to more consistent multi-view inpainting compared to independent single-view processing, referred to as a grid prior. Building upon this idea, Instant3dit~\cite{barda2025instant3dit} further introduces a masked object dataset and trains a Stable Diffusion inpainting model~\cite{lugmayr2022repaint} to generate consistent $2 \times 2$ grid images. The resulting multi-view images are then fed into a Large Reconstruction Model (LRM)~\cite{hong2023lrm} to efficiently produce a full 3D reconstruction. Despite these efforts toward cross-view consistency, such approaches are inherently limited to enforcing consistency over only a small number of views (e.g., four), which restricts viewpoint coverage and leads to insufficient geometric and textural details.  Consequently, they are mainly effective for objects with simple structures and relatively uniform textures, while struggling with more complex shapes and dense-view settings, where high-quality reconstruction remains challenging. Simply increasing the grid size leads to a trade-off between spatial coverage and resolution, resulting in degraded reconstruction quality.

To address the limitations of sparse-view 3D object editing, we propose to model 3D inpainting as a dense-view sequence problem and introduce a new framework that explicitly captures cross-view dependencies. Unlike prior approaches that process views independently or rely on sparse observations, our method jointly optimizes a sequence of densely sampled views along a $360^\circ$ trajectory, enabling globally consistent reconstruction. A key insight of our approach is that achieving high-quality 3D inpainting requires simultaneously modeling appearance generation, region completion, and  object-level structural consistency. Motivated by video generation models \cite{jiang2025vace}, we design a new 3D-object-centric architecture, tri-encoder, composed of three complementary components: a \emph{Temporal-Driven Generative Encoder} that models dependencies across densely sampled views, a \emph{Semantic-Aware Completion Encoder} that infers plausible content for missing regions, and a \emph{Cycle-Consistent 3D Encoder} that enforces global coherence through a closed-loop formulation. By duplicating the first view as the last, we explicitly impose cyclic constraints that enable long-range consistency across the entire view sequence. Furthermore, our framework naturally supports reference-guided 3D inpainting by introducing a reference view with an all-zero mask, which serves as an appearance anchor while preserving geometric consistency across views. Extensive experiments demonstrate that our method significantly outperforms prior approaches in both reconstruction quality and efficiency. In particular, ObjFiller-3D achieves more faithful and fine-grained 3D reconstructions, surpassing NeRFiller and Instant3dit with a PSNR of 26.6 (vs.\ 15.9) and an LPIPS of 0.19 (vs.\ 0.25), respectively.


Overall, our main contributions can be summarized as follows:
(1) We introduce a cycle-consistent dense-view formulation for 3D object inpainting that enforces long-range coherence across $360^\circ$ viewpoints, and instantiate it with a tri-encoder framework that jointly models dense-view generation, region completion, and global 3D consistency.
(2) We develop a reference-guided inpainting mechanism that enables controllable appearance generation while preserving structural consistency. 
(3) We achieve state-of-the-art performance on multiple benchmarks, significantly improving both reconstruction fidelity and efficiency over prior methods.


\section{Related Work}

3D object inpainting is challenging due to the need for cross-view consistency under incomplete observations. Unlike 2D inpainting, inconsistencies across viewpoints can accumulate and degrade the reconstructed 3D results. Our problem is related to several research directions, including 3D object generation and 3D editing. We review these directions below and discuss their limitations for dense-view consistent 3D inpainting.

\begin{figure*}[t]
  \centering
  \includegraphics[width=\textwidth]{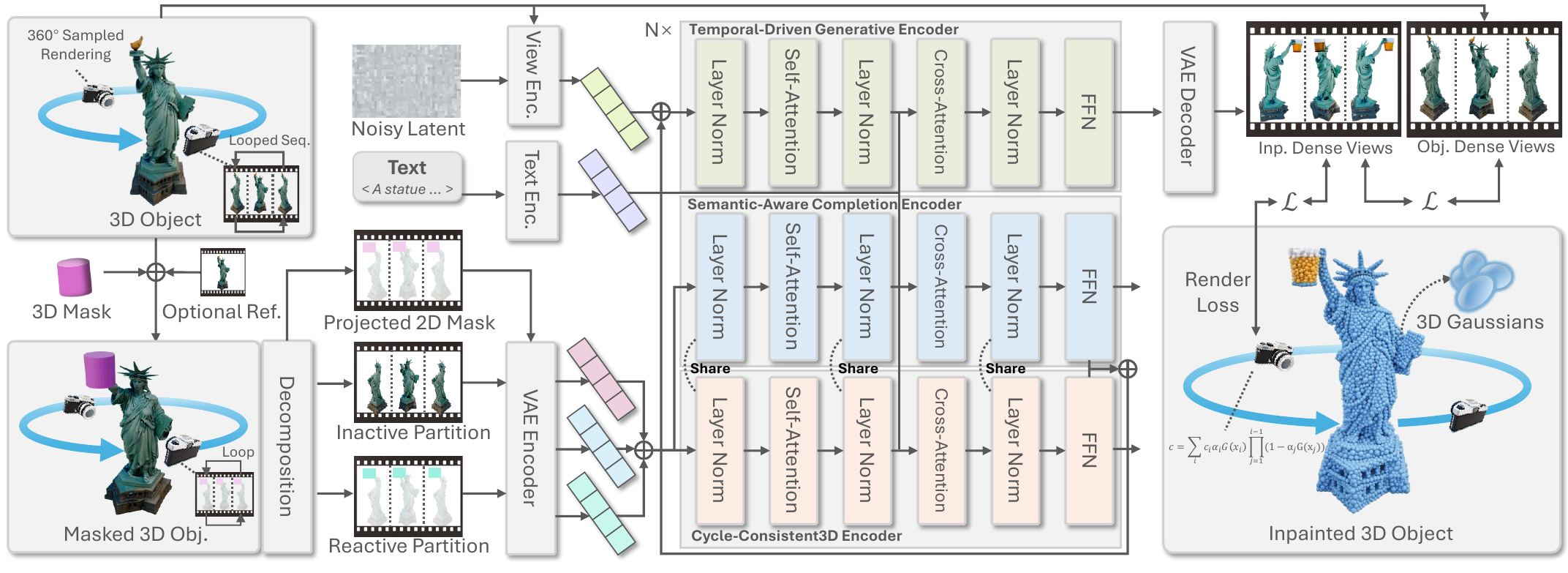}
  \caption{\textbf{Overview of ObjFiller-3D.} A complete 3D object is rendered into a looped dense-view target sequence, while a 3D mask is applied to construct a masked 3D input. The masked dense views are decomposed into inactive and reactive partitions, encoded together with the text prompt and an optional reference image, and processed by the Temporal-Driven Generative Encoder, Semantic-Aware Completion Encoder, and Cycle-Consistent 3D Encoder. The predicted latent completions are decoded into globally consistent inpainted dense views and subsequently reconstructed as a complete 3D Gaussian object. }
  \Description{VACE model architecture.}
  \label{fig:framework}
\end{figure*}

\textit{3D Object Generation.}
3D object generation methods are typically categorized into 2D prior-based and feedforward approaches. The former utilizes image generative models (e.g., Stable Diffusion~\cite{rombach2022high}, Imagen~\cite{saharia2022photorealistic}) trained on large-scale datasets like LAION-400M/5B~\cite{schuhmann2021laion,schuhmann2022laion} to extend text or image inputs into 3D. DreamFusion~\cite{poole2022dreamfusion} introduces Score Distillation Sampling (SDS) for this purpose, followed by improved but costly variants~\cite{huang2023dreamtime,wang2023prolificdreamer}. 
Alternatively, methods such as MVDream~\cite{shi2023mvdream}, ImageDream~\cite{wang2023imagedream}, Zero123++~\cite{shi2023zero123++}, SyncDreamer~\cite{liu2023syncdreamer}, Wonder3D~\cite{long2024wonder3d}, Cat3D~\cite{gao2024cat3d}, and MV-Adapter~\cite{huang2024mv} directly generate multi-view consistent images using finetuned diffusion models, offering a more efficient path to 3D reconstruction. Feedforward generation is another key paradigm for 3D synthesis, often employing encoder-decoder architectures such as Latent Diffusion Models (LDMs)~\cite{rombach2022high} to extract image tokens and generate latent 3D representations. Models such as Large Reconstruction Model (LRM)~\cite{hong2023lrm}, MeshLRM~\cite{wei2024meshlrm}, Instant3D~\cite{li2023instant3d}, and InstantMesh~\cite{xu2024instantmesh} use Transformers to map tokens into implicit triplane representations. CLAY~\cite{zhang2024clay} and CraftsMan3D~\cite{li2024craftsman3d} reconstruct meshes via neural signed distance or occupancy fields, while Trellis~\cite{xiang2025structured} and Direct3D-S2~\cite{wu2025direct3d} adopt voxel grids for more interpretable latent spaces. 
However, these methods focus on generation rather than completing partially observed objects, making them less suitable for 3D inpainting tasks.

\textit{3D Editing.}
Recent advances in 3D field editing mainly leverage NeRF~\cite{mildenhall2021nerf} and 3DGS~\cite{kerbl20233d} for superior realism and flexibility over traditional mesh or point clouds. Instruction-guided methods use pre-trained 2D diffusion models due to scarce 3D data, optimizing NeRF via Score Distillation Sampling (SDS) to achieve view-consistent, relightable 3D outputs. Iterative Dataset Update (IDU) approaches such as Instruct-NeRF2NeRF~\cite{haque2023instruct} train 3D scenes using edited images. Methods such as VcEdit~\cite{wang2024view}, GaussianEditor~\cite{chen2024gaussianeditor}, and ProEdit~\cite{chen2024proedit} extend 3DGS for efficient, controllable editing, with DGE~\cite{chen2024dge} improving multi-view coherence via epipolar attention. For 3D object removal, SPIn-NeRF~\cite{mirzaei2023spin} combines depth and 2D inpainting but struggles with full $360^\circ$ views. Recent Gaussian Splatting inpainting works such as InFusion~\cite{liu2024infusion}, AuraFusion360~\cite{wu2025aurafusion360}, and IMFine~\cite{shi2025imfine} address $360^\circ$ unbounded scene restoration. 
However, these methods focus on scene-level editing or object removal, and are not designed for consistent completion of partially observed 3D objects.

\textit{Among the most closely related methods to ours} are NeRFiller~\cite{weber2024nerfiller}, Instant3dit~\cite{barda2025instant3dit}, and Fillerbuster~\cite{weber2025fillerbuster}. Both NeRFiller and Instant3dit follow sparse-view editing paradigms, where consistency is enforced over only a limited number of input views, and NeRFiller further relies on a time-consuming IDU process. These limitations inherently restrict viewpoint coverage and hinder efficient modeling, leading to incomplete geometry and degraded texture fidelity in the reconstructed 3D objects. 
Fillerbuster does not leverage strong video priors but instead trains a 1B DiT from scratch. It uses camera ray embeddings to provide geometric information for multi-view inputs and processes all views simultaneously, which leads to increasing cross-view inconsistencies as the number of views grows.

\section{Method}

We present \textit{ObjFiller-3D}, a new framework for consistent and controllable 3D object inpainting. Our key idea is to reformulate 3D inpainting as a dense-view modeling problem, where a sequence of views sampled along a $360^\circ$ trajectory is jointly optimized to enforce global coherence across viewpoints. Given a partially observed 3D object, we construct dense-view inputs with corresponding masks, text conditions, and an optional reference image. Our model predicts a set of globally consistent inpainted views, which are subsequently used to reconstruct the completed 3D object. Motivated by video generation models \cite{jiang2025vace}, we introduce a novel 3D object-centric framework based on a tri-encoder architecture, which explicitly captures cross-view dependencies while enabling semantic-aware completion and global structural consistency. In the following, we first introduce the dense-view formulation and conditional representation, and then present the design of the three encoders and the overall optimization process.

\subsection{Preliminary and Problem Formulation}

\textit{Preliminary.}
3D Gaussian Splatting (3DGS) \cite{kerbl20233d} uses 3D Gaussians as scene primitives and represents the scene with a large number of 3D Gaussians. A Gaussian ellipsoid centered at point $\mu \in \mathbb{R}^{3}$ with covariance matrix $\Sigma$ can be written as
\begin{equation}
G(\mathrm{x})=e^{-\frac{1}{2}\mathrm{x}^T\Sigma^{-1}\mathrm{x}},\quad\text{and}\quad\Sigma = RSS^TR^T,
\end{equation}
where $\mathrm{x}$ denotes the distance from a specific point in space to the mean $\mu$, and $R \in \mathbb{R}^{3 \times 3}$ and $S \in \mathbb{R}^{3 \times 3}$ represent the Gaussian's rotation and scaling matrices, respectively. After formalizing the 3D Gaussians, volumetric rendering techniques \cite{kajiya1984ray} can be used to splat these Gaussians onto the 2D image plane. Specifically, the color $c$ for a pixel along a ray is given by
\begin{equation}
c = \sum_{i} c_i \alpha_i G(\mathrm{x_i}) \prod_{j=1}^{i-1} \left(1 - \alpha_j G(\mathrm{x_j})\right),
\end{equation}
where $\alpha \in \mathbb{R}$ and $c \in \mathbb{R}^3$ represent the opacity and color of the 3D Gaussians, respectively. 3DGS is recognized for its efficiency in real-time radiance field rendering and has thus been integrated into our method.


\textit{Problem Formulation.}
We define the task of 3D object inpainting as follows: given an incomplete 3D object $O$, a 3D mask region $M$, a text description $y$ of the object, and an optional reference image $I_r$, our goal is to generate a plausible shape $O'$ within the masked region $M$, conditioned on $y$ and $I_r$, such that the completed object $\{O \cup O'\}$ forms a geometrically consistent and visually coherent 3D structure. This task is inherently challenging, as it requires jointly reasoning about geometry, appearance, and semantic consistency under partial observations. The generated content $O'$ must not only align with the existing structure of $O$, but also maintain consistency across viewpoints when rendered from different camera poses. In addition, the completion should respect both the semantic guidance provided by the text description and the appearance cues from the reference image, if available.

\subsection{Dataset Preparation}
\begin{figure}[t]
  \centering
  \includegraphics[width=\linewidth]{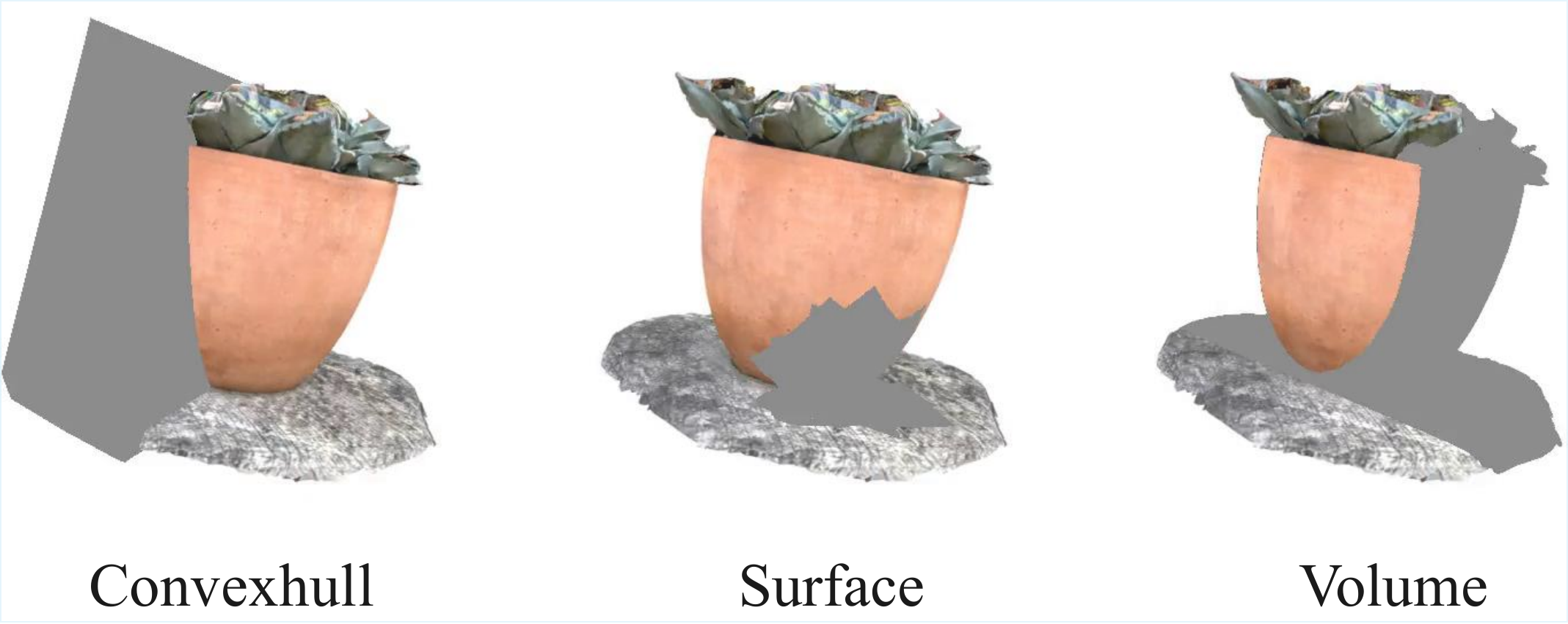}
  \caption{Three types of 3D masks.}
  \label{fig:mask_type}
  \Description{Three types of 3D masks.}
\end{figure}

We adopt the Instant3dit dataset~\cite{barda2025instant3dit}, which contains \(\sim\)7k high-quality 3D objects from Objaverse \cite{deitke2023objaverse} and three mask types: \textit{convexhull}, \textit{surface}, and \textit{volume} (Fig.~\ref{fig:mask_type}). \textit{1) Convexhull}: the missing part \(O'\) lies inside mask \(M\). \textit{2) Surface}: \(M\) covers a local surface region of object \(O\). \textit{3) Volume}: \(M\) tightly encloses the entire object \(O\).

\begin{figure}[t]
  \centering
  \includegraphics[width=\linewidth]{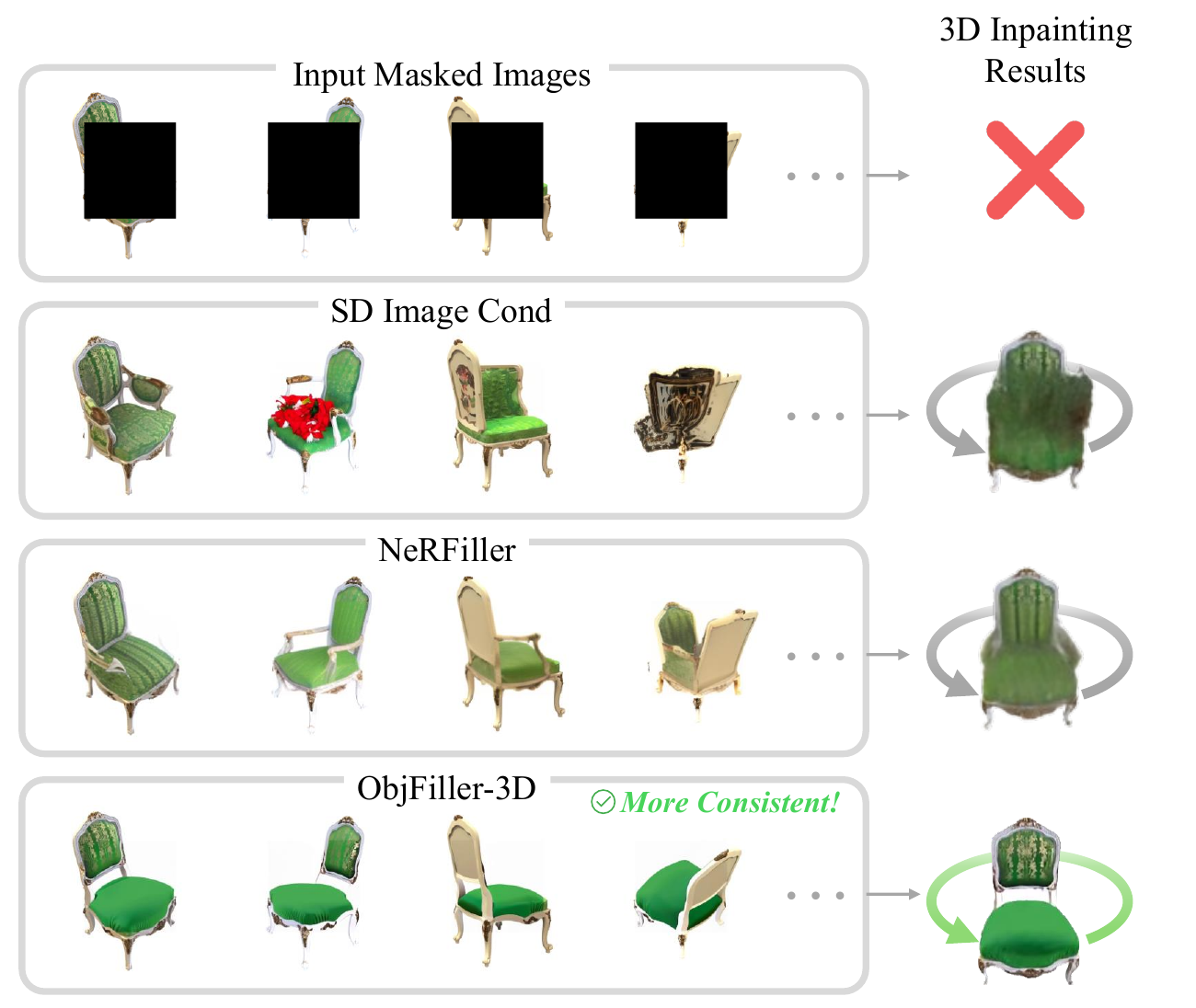}
  \caption{Visual comparison of different multi-view inpainting methods. Compared to baseline approaches, ObjFiller-3D achieves more consistent and coherent inpainting across views.}
  \Description{Visual comparison of different multi-view inpainting methods.}
  \label{fig:nerfiller}
\end{figure}

Since Instant3dit does not provide complete objects or 2D ground-truth images, we retrieve the original objects from Objaverse and render 16-view images at \(512 \times 512\) resolution using Blender. The views are sampled with a fixed 20° elevation and uniformly distributed azimuths from \(0\) to \(2\pi\). We further employ Cap3D \cite{luo2023scalable} to generate captions using BLIP-2 \cite{li2023blip}, CLIP \cite{radford2021learning}, and GPT4 \cite{achiam2023gpt}.

To exploit the inpainting prior of generative video diffusion models, we project 3D objects into 2D sequences. Specifically, object \(O\) and mask \(M\) are rendered into image frames \(F_s=\{f_i\}_{i=1}^n\) and binary masks \(M_s=\{m_i\}_{i=1}^n\) under predefined camera poses \(\Pi=\{\pi_i\}_{i=1}^{n}\):
\begin{equation}
\langle f_i, m_i \rangle = \mathcal{R}(O, M, \pi_i), \quad i \in \{1, \ldots, n\},
\label{eq:projection}
\end{equation}
where \(\mathcal{R}\) denotes the rendering operator. The resulting dataset is approximately 18 GB and will be publicly released.

\subsection{Dense-view Consistent 3D Object Inpainting}
To overcome the limitations of sparse-view editing, where each view is processed largely independently and often leads to cross-view inconsistencies, we adopt a dense-view formulation to explicitly model global coherence across viewpoints. Our method jointly optimizes a sequence of densely sampled views arranged along a $360^\circ$ trajectory, enabling consistent 3D inpainting with improved geometric and appearance fidelity. 
Let $F_s = \{f_i\}_{i=1}^{K}$ denote the rendered view sequence of an object, where each $f_i$ corresponds to a viewpoint sampled along a $360^\circ$ trajectory and $K$ is the number of dense views. Let $M_s = \{m_i\}_{i=1}^{K}$ denote the corresponding binary masks indicating regions to be inpainted, and let $y$ denote the text condition for semantic guidance. An optional reference image $I_r$ provides additional appearance cues. Each object is thus represented as a tuple $\langle F_s, M_s, y, I_r \rangle$.

To enforce cyclic consistency, we duplicate the first view and its mask as the $(K+1)$-th element, forming a closed-loop sequence. Based on this dense-looped representation, our model consists of three core components: a \emph{Temporal-Driven Generative Encoder}, a \emph{Semantic-Aware Completion Encoder}, and a \emph{Cycle-Consistent 3D Encoder}. The first branch captures long-range dependencies over the entire dense-view loop from noisy latent views, the second branch performs text- and mask-guided completion on the conditional latent sequence, and the third branch further refines the completion stream to enforce cross-view cycle consistency. Each of these components is implemented as a stack of multiple encoder layers. Together, they transform the input dense-view sequence into the inpainted sequence $F_s'$, which remains globally consistent across viewpoints and is subsequently used for 3DGS reconstruction.

\textit{Conditional Embedding.}
Inspired by the pretrained VACE\cite{jiang2025vace} input pipeline, we reuse three inherited front-end modules in Fig.~\ref{fig:framework}: a view encoder, a text encoder, and a VAE encoder/decoder. 
The \emph{view encoder} maps the looped dense-view target sequence rendered from the complete object into latent video tokens, which are further perturbed with Gaussian noise to form the noisy latent $x_t$ used by the generative branch. 
The \emph{text encoder} transforms the prompt $y$ into semantic tokens $c_y$, which are injected into the transformer blocks through cross-attention. 
The \emph{VAE encoder} projects the masked conditional inputs, including the inactive partition, the reactive partition, the projected 2D masks, and the optional reference image, into a shared latent space, while the inherited \emph{VAE decoder} finally maps the predicted latent sequence back to dense-view images. 
Specifically, during training, we first render the complete 3D object into a looped dense-view supervision sequence 
\(F_s^{\mathrm{obj}} = \{f_i^{\mathrm{obj}}\}_{i=1}^{K+1}\), where the first view is duplicated as the \((K+1)\)-th frame to enforce cyclic consistency. We then apply a 3D mask to obtain the masked dense-view input \(F_s = \{f_i\}_{i=1}^{K+1}\) and masks \(M_s = \{m_i\}_{i=1}^{K+1}\).
Each masked view is decomposed into inactive and reactive partitions:
\begin{equation}
f_i^{\mathrm{inactive}} = f_i \odot (1-m_i), \qquad
f_i^{\mathrm{reactive}} = f_i \odot m_i.
\end{equation}

The VAE encoder maps them into latent representations:
\begin{equation}
z_i = \mathrm{Concat}\!\left(
\phi(f_i^{\mathrm{inactive}}),
\phi(f_i^{\mathrm{reactive}}),
\tilde{m}_i
\right),
\end{equation}
where \(\phi(\cdot)\) denotes the inherited VAE encoder and \(\tilde{m}_i\) is the projected latent-aligned mask. If an optional reference image \(I_r\) is provided, it is encoded as \(z_r = \phi(I_r)\) and prepended temporally:
\begin{equation}
Z = \mathrm{Concat}(z_r, z_1, z_2, \dots, z_K).
\end{equation}

After transformer-based dense-view completion, the final latent sequence \(Z_{\mathrm{final}}\) is decoded by the inherited VAE decoder $F_s' = \psi(Z_{\mathrm{final}})$, where \(\psi(\cdot)\) denotes the VAE decoder and \(F_s'\) is the predicted inpainted dense-view sequence. With conditional embedding, the model operates on dense-view latent representations through a unified encoder module composed of three branches. 

The \emph{Temporal-Driven Generative Encoder} is responsible for modeling the global dependency of the looped dense-view sequence. 
It takes the noisy dense-view latent $x_t$ as input, following the inherited VACE generative path, and propagates information over the entire view loop rather than over isolated viewpoints. 
This design is particularly important for our setting, since dense-view object completion requires the model to preserve appearance and structure under large viewpoint changes, while simultaneously maintaining first-last view consistency in the closed loop. 
The output of this branch is a sequence of generative dense-view features that provides the global completion prior.

The \emph{Semantic-Aware Completion Encoder} focuses on the masked-region recovery conditioned on the observed object content. 
Its input is the conditional latent sequence $Z$, which is constructed from the inactive partition, the reactive partition, the projected 2D masks, and the optional reference image. 
Unlike the generative branch that mainly captures global dense-view dynamics, this branch explicitly emphasizes which regions should be preserved and which regions should be regenerated. 
Through cross-attention to the text tokens, it further injects semantic guidance into the masked-region completion process, allowing the model to recover missing object parts in a way that is both structurally plausible and semantically aligned with the prompt. 
The output of this branch is a sequence of completion-oriented dense-view features.

The \emph{Cycle-Consistent 3D Encoder} is designed to inject closed-loop multi-view consistency into the completion stream. 
Rather than predicting an independent dense-view sequence, it operates on the corresponding completion features and estimates cycle-consistency residuals that are fused back into the main completion branch. 
This enables the model to explicitly refine cross-view coherence in transformer space, especially under large pose variation and loop closure, where purely video-driven completion may still exhibit subtle view inconsistency. 
In practice, this encoder is realized by a factorized residual parameterization attached to the corresponding completion blocks, such that cycle-consistent 3D cues can be introduced without changing the inherited front-end encoders or the main backbone.

Together, these branches form the core transformer-based representation module of ObjFiller-3D, where global dense-view generation, semantics-aware completion, and cycle-consistent 3D refinement are jointly modeled before decoding the final inpainted views. As shown in Fig.~\ref{fig:framework}, the above three encoders are instantiated on top of the pretrained VACE backbone and share the same DiT-style block topology.
The residual features produced by the Cycle-Consistent 3D Encoder are fused into the completion stream, and the final latent sequence is decoded by the inherited VAE decoder to obtain the inpainted dense-view sequence $F_s'$.

\begin{figure}[t]
  \centering
  \includegraphics[width=\linewidth]{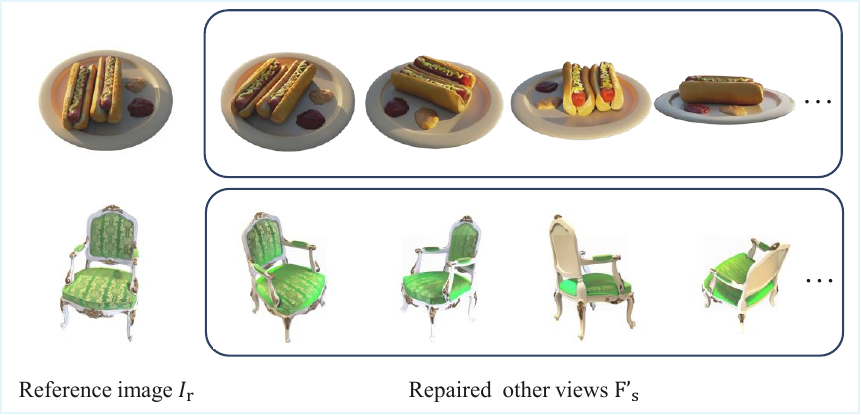}
  \caption{Reference-based 3D inpainting. Given a reference image, the generated views are well aligned with the input.}
  \label{fig:reference}
\end{figure}

\textit{Two-stage Optimization.} 
Our framework jointly addresses conditional dense-view completion and 3D novel view synthesis. We adopt a two-stage optimization scheme and train the model using a flow matching objective. Let $u_\theta\bigl(t, x \mid F_s, M_s, y\bigr)$ denote the predicted velocity field, and let $u^*(t,x)$ denote the target velocity induced by interpolating between the noise and data distributions. The loss is defined as:
\begin{equation}
\mathcal{L}_{\mathrm{FM}}(\theta)=
\mathbb{E}_{t \sim \mathcal{U}(0,1)}\, \mathbb{E}_{x \sim p_t}
\Big\|
u_\theta(t, x \mid F_s, M_s, y)
- u^*(t, x)
\Big\|_2^2,
\end{equation}
where $p_t$ denotes the marginal distribution at time $t$. Minimizing this loss encourages a smooth and consistent transformation from noise to data, yielding coherent inpainted sequences. After inpainting, we obtain completed frames $F'_s$ and corresponding camera poses $\Pi$. These are used to reconstruct a 3D object represented by Gaussian primitives $\mathcal{G}$:
\begin{equation}
\mathcal{G}_{\text{inpainted}} = \arg\min_\mathcal{G} \sum_{\pi_i \in \Pi, f'_i \in F'_s} \mathcal{L}\left(\mathcal{R}(\mathcal{G}, \pi_i), f'_i\right),
\end{equation}
where $\mathcal{R}$ denotes the differentiable renderer and $\mathcal{L}$ is the 3DGS reconstruction loss. Compared to NeRFiller, our method benefits from strong cross-view coherence established during inpainting, eliminating the need for iterative dataset update (IDU). This significantly reduces reconstruction time: our method completes the pipeline in under 10 minutes, while NeRFiller~\cite{weber2024nerfiller} requires over 40 minutes. Furthermore, our approach achieves substantially higher reconstruction quality, as shown in Figs.~\ref{fig:demo1} and~\ref{fig:nerfiller}.

\section{Experiments}

\textit{Implementation Details.}
We train our models using 3,000 objects from the reprocessed dataset. 
In our implementation, We optimize only the factorized residual branch of the Cycle-Consistent 3D Encoder while keeping the inherited VACE backbone fixed. The Cycle-Consistent 3D Encoder is instantiated by attaching factorized residual parameters to the corresponding completion blocks.
We set the factorization rank of the Cycle-Consistent 3D Encoder to 32, use a learning rate of $10^{-4}$, a batch size of 4, and train for 10 epochs. 
For the 14B backbone, training is conducted in half precision, consuming approximately 60 GB of VRAM, and takes around 3 days on a single NVIDIA A800 GPU. 
During inference, we use the UniPC sampler \cite{zhao2023unipc} with 20 sampling steps, a CFG\cite{ho2022classifier}guidance scale of 4, and a residual branch scale of 1.

\textit{Evaluation Metrics.}
For the comparison between our method and Instant3dit, we focus on three aspects. \textit{1) Text-image similarity}: we measure the semantic relevance between the generated $2 \times 2$ grid images and the given text prompt using the CLIP \cite{radford2021learning} similarity score. \textit{2) Image quality and fidelity}: this evaluates the quality of the generated images and the similarity between the inpainted images and the ground-truth (GT) images. Specifically, we inpaint 300 grid images from the evaluation dataset and compare them against GT grids using the FID score~\cite{Heusel_Ramsauer_Unterthiner_Nessler_Hochreiter_2017}. \textit{3) Multi-view consistency}: to assess consistency across the four views in the inpainted grid image, we feed the four images directly into LRM~\cite{xu2024instantmesh}. If the views lack consistency, LRM will produce a distorted object; otherwise, the reconstruction will be coherent. To quantify this, we render the resulting object from the original viewpoints and compute the perceptual LPIPS score~\cite{Zhang_Isola_Efros_Shechtman_Wang_2018} between the rendered images and the inpainted ones. A lower LPIPS score indicates better multi-view consistency. For comparison with NeRFiller and Fillerbuster, the final object reconstruction is a 3D reconstruction task. Therefore, we evaluate using three standard metrics: SSIM, PSNR, and LPIPS.

\begin{figure}[t]
  \centering
  \includegraphics[width=\linewidth]{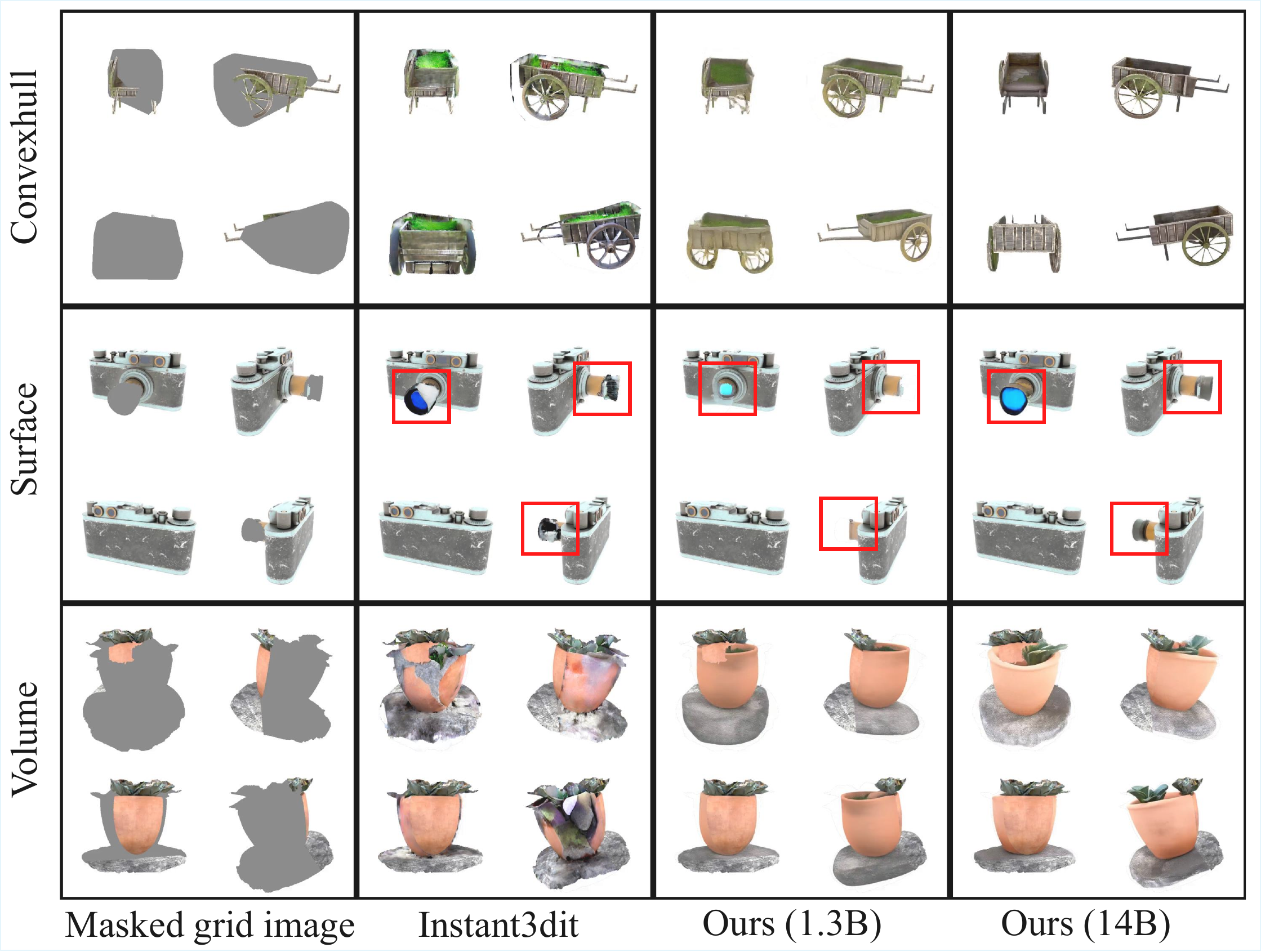}
  \caption{Qualitative results compared with Instant3dit. We apply our method to three different mask types, listed from top to bottom as Convexhull, Surface, and Volume. ObjFiller-3D-14B (the rightmost column) exhibits the highest consistency across all mask types.}
  \Description{Qualitative results compared with Instant3dit.}
  \label{fig:instant3dit}
\end{figure}

\begin{table*}[h]
\caption{\textbf{Quantitative comparison on NeRF Blender and NeRFiller datasets.} Cells are highlighted as follows: \colorbox{red!40}{best}, \colorbox{orange!40}{second best}.}
\centering
\scalebox{0.9}{
\begin{tabular}{l|ccc|ccc}
\toprule
\multirow{2}{*}{Method} & \multicolumn{3}{c|}{NeRF Blender \cite{mildenhall2021nerf}} & \multicolumn{3}{c}{NeRFiller \cite{weber2024nerfiller}} \\
 & PSNR $\uparrow$ & SSIM $\uparrow$ & LPIPS $\downarrow$ & PSNR $\uparrow$ & SSIM $\uparrow$ & LPIPS $\downarrow$ \\
\midrule
GT & 35.65 & 0.97 & 0.03 & 35.22 & 0.98 & 0.02 \\
SD Image Cond & 14.15 & 0.76 & 0.28 & 19.89 & 0.89 & 0.12 \\
NeRFiller (CVPR’24) \cite{weber2024nerfiller} & 15.89 & 0.82 & 0.23 & 28.85 & 0.95 & 0.04 \\
Fillerbuster (3DV'26) \cite{weber2025fillerbuster} & 20.43 & 0.85 & 0.21 & \cellcolor{orange!40}{32.82} & \cellcolor{red!40}{0.97} & \cellcolor{orange!40}{0.06} \\
\textbf{ObjFiller-3D-1.3B (Ours)} & \cellcolor{orange!40}{21.38} & \cellcolor{orange!40}{0.88} & \cellcolor{orange!40}{0.11} & 30.75 & 0.96 & \cellcolor{orange!40}{0.06} \\
\textbf{ObjFiller-3D-14B (Ours)} & \cellcolor{red!40}{26.62} & \cellcolor{red!40}{0.93} & \cellcolor{red!40}{0.07} & \cellcolor{red!40}{33.68} & \cellcolor{red!40}{0.97} & \cellcolor{red!40}{0.03} \\
\bottomrule
\end{tabular}}
\label{tab:comparison_nerf*}
\end{table*}

\begin{table}[h]
\caption{\textbf{Quantitative comparison on Instant3dit dataset.} Cells are highlighted as follows: \colorbox{red!40}{best}, \colorbox{orange!40}{second best}.}
\centering
\scalebox{0.9}{
\begin{tabular}{l|ccc}
\toprule
\multirow{2}{*}{Method} & \multicolumn{3}{c}{Instant3dit \cite{barda2025instant3dit}} \\
 & FID $\downarrow$ & LPIPS $\downarrow$ & Clip $\uparrow$ \\
\midrule
GT & - & 0.140 & 30.53 \\
SD Image Cond & 120.5 & 0.277 & 28.73 \\
Instant3dit (CVPR'25) \cite{barda2025instant3dit} & 100.9 & 0.253 & 29.81 \\
Fillerbuster (3DV'26) \cite{weber2025fillerbuster} & 107.4 & 0.259 & 29.23 \\
\textbf{ObjFiller-3D-1.3B (Ours)} & \cellcolor{orange!40}{92.07} & \cellcolor{red!40}{0.190} & \cellcolor{orange!40}{29.87} \\
\textbf{ObjFiller-3D-14B (Ours)} & \cellcolor{red!40}{90.75} & \cellcolor{orange!40}{0.195} & \cellcolor{red!40}{30.19} \\
\bottomrule
\end{tabular}}
\label{tab:comparison_instant3dit}
\end{table}

\subsection{Comparisons with State-of-the-Arts}

\textit{Comparisons on Instant3dit Dataset.}
We consider four variants: the vanilla VACE-1.3B and VACE-14B backbones, and their counterparts equipped with the proposed Cycle-Consistent 3D Encoder. 
We evaluate all variants on 300 held-out objects. 
For Instant3dit, we stack masked multi-view images and masks from four orthogonal viewpoints ($0^\circ$, $90^\circ$, $180^\circ$, and $270^\circ$) into $2 \times 2$ grids. These image grids, along with text prompts generated by Cap3D \cite{luo2023scalable}, serve as input to Instant3dit's open-source 2D multi-view inpainting model, which has been trained on SDXL \cite{podell2023sdxl}, exactly as expected by its design. 
For our method, we concatenate all dense views and projected masks into 17-frame looped sequence, and feed them into the inherited VACE pipeline together with text prompts. 
From the predicted inpainted dense-view sequence, we extract the same four orthogonal views and arrange them into $2 \times 2$ grids for evaluation. 
Since object reconstruction in later steps is uniformly performed by LRM for all methods, this comparison focuses on the quality, consistency, and semantic fidelity of the generated dense-view images.

\textit{Comparisons on NeRF Blender and NeRFiller Dataset.}
We choose ObjFiller-3D-14B, i.e., the 14B VACE backbone equipped with the proposed Cycle-Consistent 3D Encoder and followed by 3DGS reconstruction, as our main model for comparison with NeRFiller and Fillerbuster. 
For a more comprehensive evaluation, we also include SD Image Cond, where each view is individually inpainted by Stable Diffusion and directly used for reconstruction. 
Experiments are conducted on the NeRF Blender synthetic dataset \cite{mildenhall2021nerf}, with a $256 \times 256$ occlusion at the image center (see Figure \ref{fig:nerfiller}, top), and on the NeRFiller dataset \cite{weber2024nerfiller}, which contains scanned meshes with corresponding masks.

\subsection{Further Analysis}

The quantitative results are summarized in Table~\ref{tab:comparison_nerf*} and Table~\ref{tab:comparison_instant3dit}. 
We exclude Instant3dit from the NeRF Blender and NeRFiller benchmarks in Table~\ref{tab:comparison_nerf*} because Instant3dit is strictly limited to four-view inputs, making it inapplicable to scenarios with numerous sparse views. Similarly, NeRFiller is omitted from the Instant3dit dataset in Table~\ref{tab:comparison_instant3dit} as it reduces to the SD Image Cond baseline under the four-view constraint.
Overall, the 14B variant of ObjFiller-3D achieves the best performance across datasets. 
The qualitative results in Fig.~\ref{fig:instant3dit} further verify that introducing the proposed Cycle-Consistent 3D Encoder consistently improves dense-view coherence and visual fidelity.

\textit{Number of Input Views.}
More incomplete views provide extra object information, but also impose greater constraints. As shown in Table~\ref{tab:input_view}, the performance of our inpainter improves with more inputs. Therefore, we argue that our method effectively handles the negative aspects introduced by additional views, while also benefiting from the positive information they provide.

\begin{table}[t]
  \caption{Evaluation of our 3D inpainting method across different numbers of input views.}
  \label{tab:input_view}
  \centering
  \renewcommand{\arraystretch}{1.2}
  \setlength{\tabcolsep}{10pt}
  \begin{tabular}{l|cccc}
    \toprule
    Input Views & 80 & 100 & 120 & 140 \\
    \midrule
    PSNR $\uparrow$ & 22.76 & 25.63 & 26.62 & 26.68 \\
    SSIM $\uparrow$ & 0.89 & 0.92 & 0.93 & 0.93 \\
    LPIPS $\downarrow$ & 0.11 & 0.08 & 0.07 & 0.06 \\
    \bottomrule
  \end{tabular}
\end{table}

\begin{figure*}[t]
  \centering
  \includegraphics[width=0.95\textwidth]{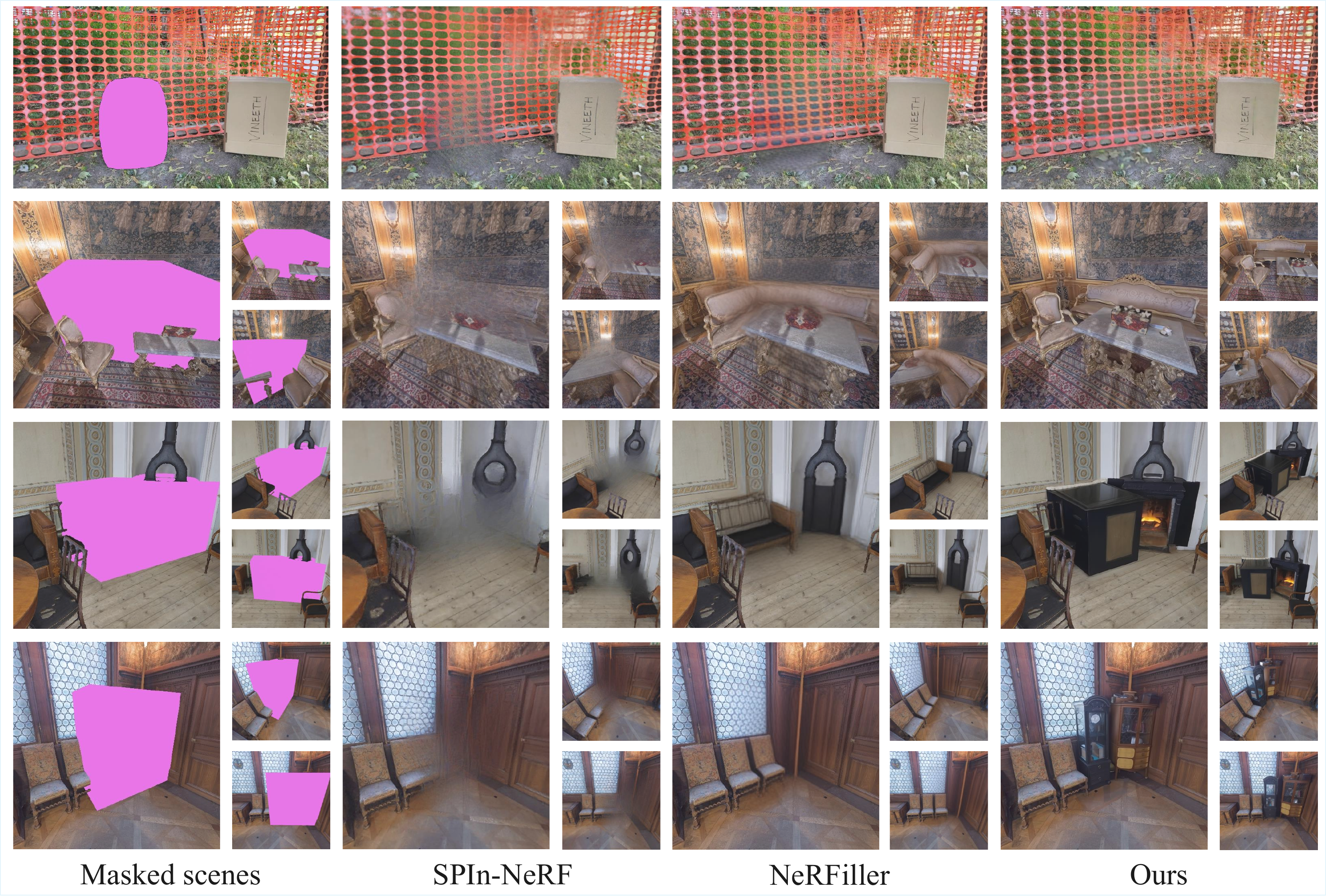}
  \caption{Qualitative results of scene inpainting. The leftmost column shows the scenes we need to process, where the areas to be repaired are masked in pink. We compare our method, ObjFiller-3D (the rightmost column), with other approaches.}
  \label{fig:scenes}
\end{figure*}

\textit{Extend to 3D Scene.}
Leveraging our dense-view-based approach, we naturally extend it to 3D scene inpainting, which, like object-level inpainting, is fundamentally a mask-filling task. We evaluate our method on four diverse scenes, comparing it with NeRFiller and SPIn-NeRF \cite{mirzaei2023spin}, as shown in Fig.~\ref{fig:scenes}. Unlike SPIn-NeRF, which assumes compact mask regions suited for object removal, our approach accommodates broader, unconstrained masks, enabling more general 3D inpainting and demonstrating greater applicability. In complex scenes, our method outperforms NeRFiller, as illustrated in the second row of Fig.~\ref{fig:scenes}.

\begin{figure*}[t]
  \centering

  \begin{subfigure}[t]{0.48\linewidth}
    \centering
    \includegraphics[width=\linewidth]{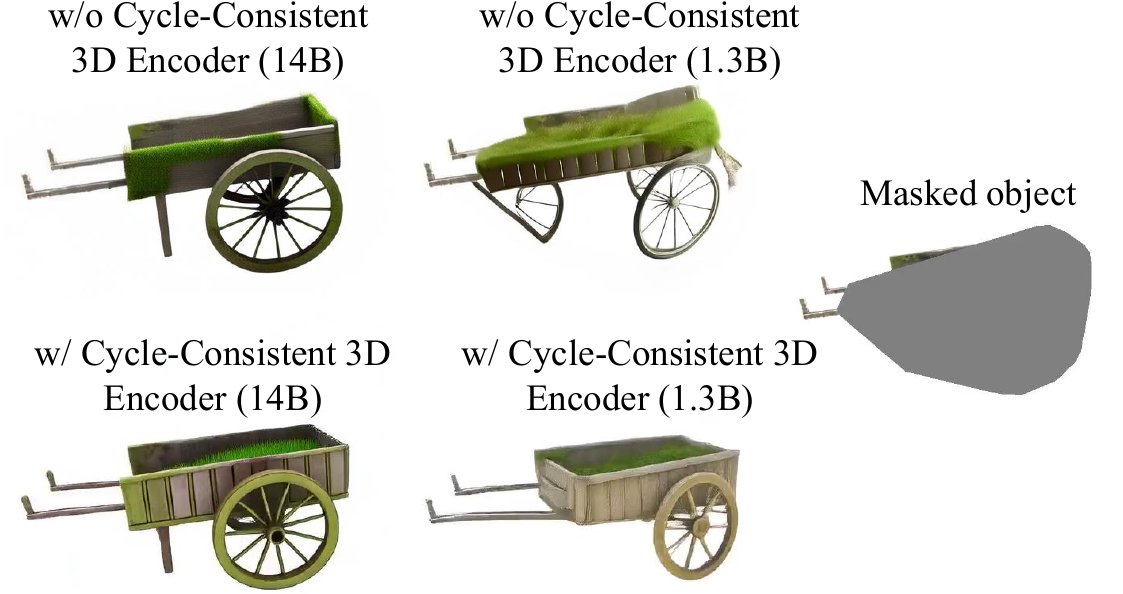}
    \caption{Cycle-Consistent 3D Encoder ablation.}
    \label{fig:Cycle-Consistent_ablation}
  \end{subfigure}
  \hfill
  \begin{subfigure}[t]{0.48\linewidth}
    \centering
    \includegraphics[width=\linewidth]{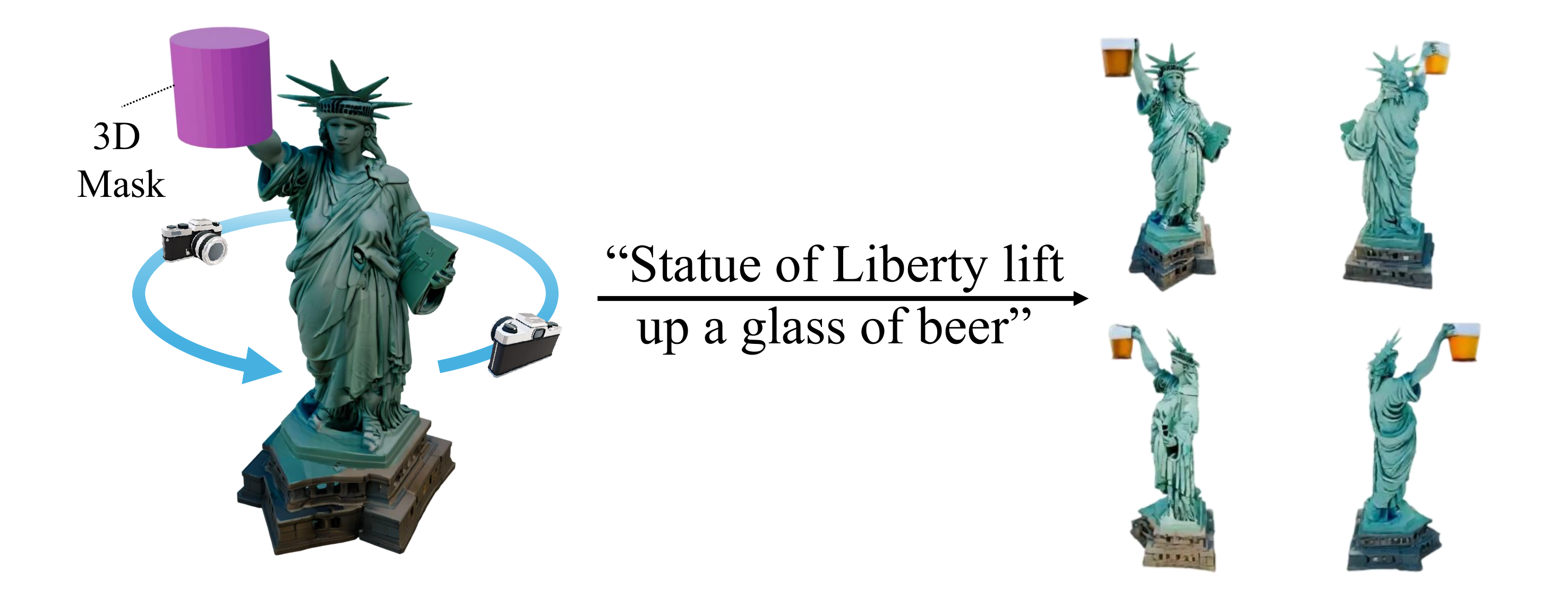}
    \caption{Object editing.}
    \label{fig:statue}
  \end{subfigure}

  \caption{Additional results and ablations of ObjFiller-3D.}
  \label{fig:combined}
\end{figure*}

\textit{Effectiveness of Cycle-Consistent 3D Encoder.}
A key component of our method is the proposed Cycle-Consistent 3D Encoder, which is trained under three types of multi-view-consistent masks. To assess their effectiveness, we conduct ablation studies focusing on this aspect. Fig.~\ref{fig:Cycle-Consistent_ablation} shows the difference. The quantitative comparisons are provided in Table~\ref{tab:cce_ablation}.

\begin{table}[t]
\centering
\caption{Ablation study of Cycle-Consistent 3D Encoder (CCE) on 1.3B and 14B models. The inclusion of CCE significantly enhances reconstruction quality and semantic consistency.}
\label{tab:cce_ablation}
\begin{tabular}{l|l|ccc}
\toprule
Model Name & Configuration & FID $\downarrow$ & LPIPS $\downarrow$ & CLIP $\uparrow$ \\
\midrule
\multirow{2}{*}{ObjFiller-3D-1.3B} & w/o CCE & 107.20 & 0.205 & 29.76 \\
 & w/ CCE  & \textbf{92.07} & \textbf{0.190} & \textbf{29.87} \\
\midrule
\multirow{2}{*}{ObjFiller-3D-14B} & w/o CCE & 104.80 & 0.219 & \textbf{30.19} \\
 & w/ CCE  & \textbf{90.75} & \textbf{0.195} & \textbf{30.19} \\
\bottomrule
\end{tabular}
\end{table}

\textit{Applications.}
Since inpainting and editing are closely related tasks, our model can also be applied to object editing. Specifically, we can import our object into Blender and manually place a 3D geometric shape as the 3D mask in the desired location. Fig.~\ref{fig:statue} illustrates a \textit{replace} example. Similarly, ObjFiller-3D supports \textit{add} and \textit{remove}.

\begin{figure*}[htbp]
  \centering
  \includegraphics[width=0.95\textwidth]{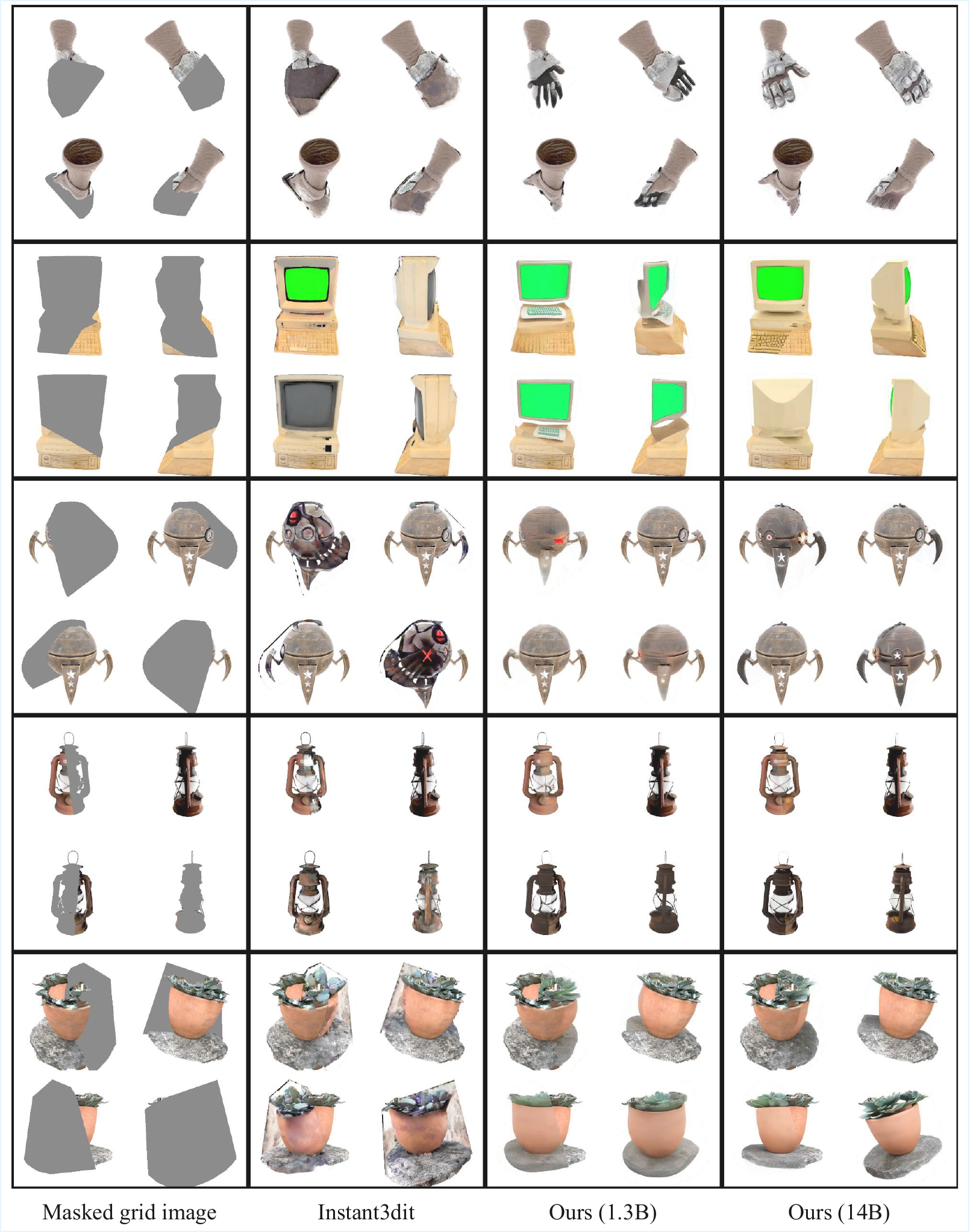}
  \caption{\textbf{Additional visualization results compared to Instant3dit.} We randomly select several outputs from our method and Instant3dit for qualitative comparison. Overall, our 14B model demonstrates superior consistency across multiple views.}
  \label{fig:additional}
\end{figure*}

\textit{Additional Results.} We randomly select several outputs from our method and Instant3dit for qualitative comparison. Overall, our 14B model demonstrates superior consistency across multiple views. Figure \ref{fig:additional} presents additional visualization results for comparative analysis with Instant3dit.

\section{Conclusion}
In this paper, we presented ObjFiller-3D, a novel framework for high-quality and consistent 3D object inpainting. Unlike prior sparse-view or per-view completion methods, our approach jointly models densely sampled views along a $360^\circ$ trajectory, enabling coherent geometry and appearance across viewpoints. Our framework combines dense-view modeling, object-level completion, and global structural consistency within a unified architecture, while also supporting reference-guided 3D inpainting for controllable appearance editing. Extensive experiments showed that ObjFiller-3D consistently outperforms existing methods in reconstruction fidelity and perceptual quality. We believe our work provides a promising framework for scalable 3D content completion and editing.

\textit{Limitation.} Our approach relies on foundation models for dense-view generation, and is therefore inherently limited by their generative and consistency capabilities. We believe future advances in foundation models will further improve the quality and robustness of our framework.


\bibliographystyle{ACM-Reference-Format}
\bibliography{main}

\end{document}